%% file: main.tex
\documentclass[]{fairmeta}
\usepackage{makecell}
\usepackage{wrapfig}
\usepackage{tabularx}
\usepackage{textcomp}
\usepackage{stfloats}
\usepackage{url}
\usepackage{verbatim}
\usepackage{titlesec}
\usepackage{tocloft}
\usepackage{adjustbox}
\usepackage{multirow}
\usepackage{pifont}
\usepackage[sc]{mathpazo}
\usepackage{tikz}
\usepackage{comment}
\usepackage{amsmath,amssymb}
\usepackage{colortbl}
\usepackage{natbib}
\usepackage{color}
\usepackage{booktabs} 
\usepackage{hyperref}
\usepackage{graphicx}
\usepackage{subcaption}
\usepackage{multirow}
\usepackage{subcaption}
\RequirePackage{xspace}
\makeatletter
\DeclareRobustCommand\onedot{\futurelet\@let@token\@onedot}
\def\@onedot{\ifx\@let@token.\else.\null\fi\xspace}
\usepackage[most]{tcolorbox}
\usepackage{xcolor}
\usepackage{array}
\usepackage{tabularx}
\usepackage{siunitx} 
\usepackage{makecell}
\usepackage[table]{xcolor}
\usepackage{caption}
\definecolor{headerpurple}{HTML}{d8d2fc}
\definecolor{rowgray}{gray}{0.95}
\usepackage[ruled, linesnumbered, noend]{algorithm2e}
\usepackage{multirow}
\usepackage{tikz}
\usepackage{xspace}

\makeatother

\definecolor{adptorange}{RGB}{248, 205, 172}
\definecolor{cmpblue}{RGB}{189, 215, 238}
\definecolor{cmpblue}{RGB}{189, 215, 238}

\definecolor{our_red}{RGB}{232,157,160}
\definecolor{our_blue}{RGB}{136,206,230}
\definecolor{our_orange}{RGB}{246,200,168}
\definecolor{our_green}{RGB}{178,211,164}

\definecolor{attn_code0}{RGB}{247,215,200}
\definecolor{attn_code1}{RGB}{238,169,139}
\definecolor{mlp_code0}{RGB}{204,201,221}
\definecolor{mlp_code1}{RGB}{102,95,153}
\definecolor{mygray}{HTML}{f0f0f0}

\definecolor{token_blue}{RGB}{84, 120, 140}

\usetikzlibrary{shapes, arrows.meta, positioning, calc, fit, backgrounds, patterns, decorations.pathreplacing}

\usepackage{pifont}
\newcommand{\cmark}{\ding{51}}
\newcommand{\xmark}{\ding{55}}

\newcommand{\methodname}{PackForcing\xspace}

\newcommand{\xmarkg}{\textcolor{gray!40}{\ding{55}}\xspace}%

\makeatletter
\newcommand{\thickhline}{%
    \noalign {\ifnum 0=`}\fi \hrule height 1pt
    \futurelet \reserved@a \@xhline
    \noalign{\vskip 2pt}
}
\makeatother

\definecolor{sinkcolor}{HTML}{4A90D9}
\definecolor{compcolor}{HTML}{E8A838}
\definecolor{recentcolor}{HTML}{5CB85C}
\definecolor{currentcolor}{HTML}{D9534F}
\definecolor{hrcolor}{HTML}{9B59B6}
\definecolor{lrcolor}{HTML}{1ABC9C}
\definecolor{darkgray}{HTML}{555555}

\usepackage{orcidlink}
\usepackage{titletoc}
\usepackage{microtype}
\usepackage{CJKutf8}
\usepackage[utf8]{inputenc}
\usepackage{multicol}
\usepackage{setspace}
\usepackage{url}
\usepackage{tabularx}
\usepackage{pgfplots}
\pgfplotsset{compat=1.18}
\usetikzlibrary{er,positioning,bayesnet}
\usepackage[inline]{enumitem}
\usepackage{siunitx}
\usepackage{tocloft}
\usepackage{listings}
\usepackage{xltabular}
\usepackage[framemethod=tikz]{mdframed}
\usepackage{todonotes}
\usepackage{fontawesome5}
\usepackage{cleveref}

\surroundwithmdframed[
  hidealllines=true,
  innerleftmargin=0pt,
  innertopmargin=0pt,
  innerbottommargin=0pt]{lstlisting}
\lstnewenvironment{response}[1][]
 {\lstset{
 columns=fullflexible,
  breakautoindent=false, breakindent=0pt, breaklines, linewidth=8cm, #1}}
 {}
\tcbuselibrary{breakable}

\newlength\savewidth

\definecolor{url_color}{RGB}{0, 0, 255}

\hypersetup{
    colorlinks=true,
    linkcolor=black,
    urlcolor=url_color,
}

\title{PackForcing: Short Video Training Suffices for Long Video Sampling and Long Context Inference}

\author[1,2,*,\S]{Xiaofeng Mao}
\author[1,3,*,\S]{Shaohao Rui}
\author[1,2]{Kaining Ying}
\author[1]{Bo Zheng}
\author[1]{Chuanhao Li}
\author[2,\dagger]{Mingmin Chi}
\author[1,3\dagger]{Kaipeng Zhang}

\affiliation[1]{Alaya Studio, Shanda AI Research Tokyo}
\affiliation[2]{Fudan University}
\affiliation[3]{Shanghai Innovation Institute}

\abstract{Autoregressive video diffusion models have demonstrated remarkable progress, yet they remain bottlenecked by intractable linear KV-cache growth, temporal repetition, and compounding errors during long-video generation. To address these challenges, we present \textbf{\methodname{}}, a unified framework that efficiently manages the generation history through a novel three-partition KV-cache strategy. Specifically, we categorize the historical context into three distinct types: (1) \textbf{Sink tokens}, which preserve early anchor frames at full resolution to maintain global semantics; (2) \textbf{Mid tokens}, which achieve a massive spatiotemporal compression (${\sim}32\times$ token reduction) via a dual-branch network fusing progressive 3D convolutions with low-resolution VAE re-encoding; and (3) \textbf{Recent tokens}, kept at full resolution to ensure local temporal coherence. To strictly bound the memory footprint without sacrificing quality, we introduce a dynamic top-$k$ context selection mechanism for the mid tokens, coupled with a continuous Temporal RoPE Adjustment that seamlessly re-aligns position gaps caused by dropped tokens with negligible overhead. Empowered by this principled hierarchical context compression, \methodname{} can generate coherent 2-minute, $832{\times}480$ videos at 16\,FPS on a single H200 GPU. It achieves a bounded KV cache of just ${\sim}\mathbf{4}$\,\textbf{GB} and enables a remarkable $\mathbf{24\times}$ temporal extrapolation ($5\,\text{s} \rightarrow 120\,\text{s}$), operating effectively either zero-shot or trained on merely 5-second clips. Extensive results on VBench demonstrate state-of-the-art temporal consistency (26.07) and dynamic degree (56.25), proving that short-video supervision is sufficient for high-quality, long-video synthesis.}

\github{\url{https://github.com/ShandaAI/PackForcing}}
\date{\today}

\makeatletter
\renewenvironment{abstract}{\setbox\z@\vbox\bgroup}{\egroup}
\makeatother

\begin{document}

\maketitle
\vspace{-1.3em}
\begin{figure*}[!h]
    \centering
    \includegraphics[width=0.86\linewidth]{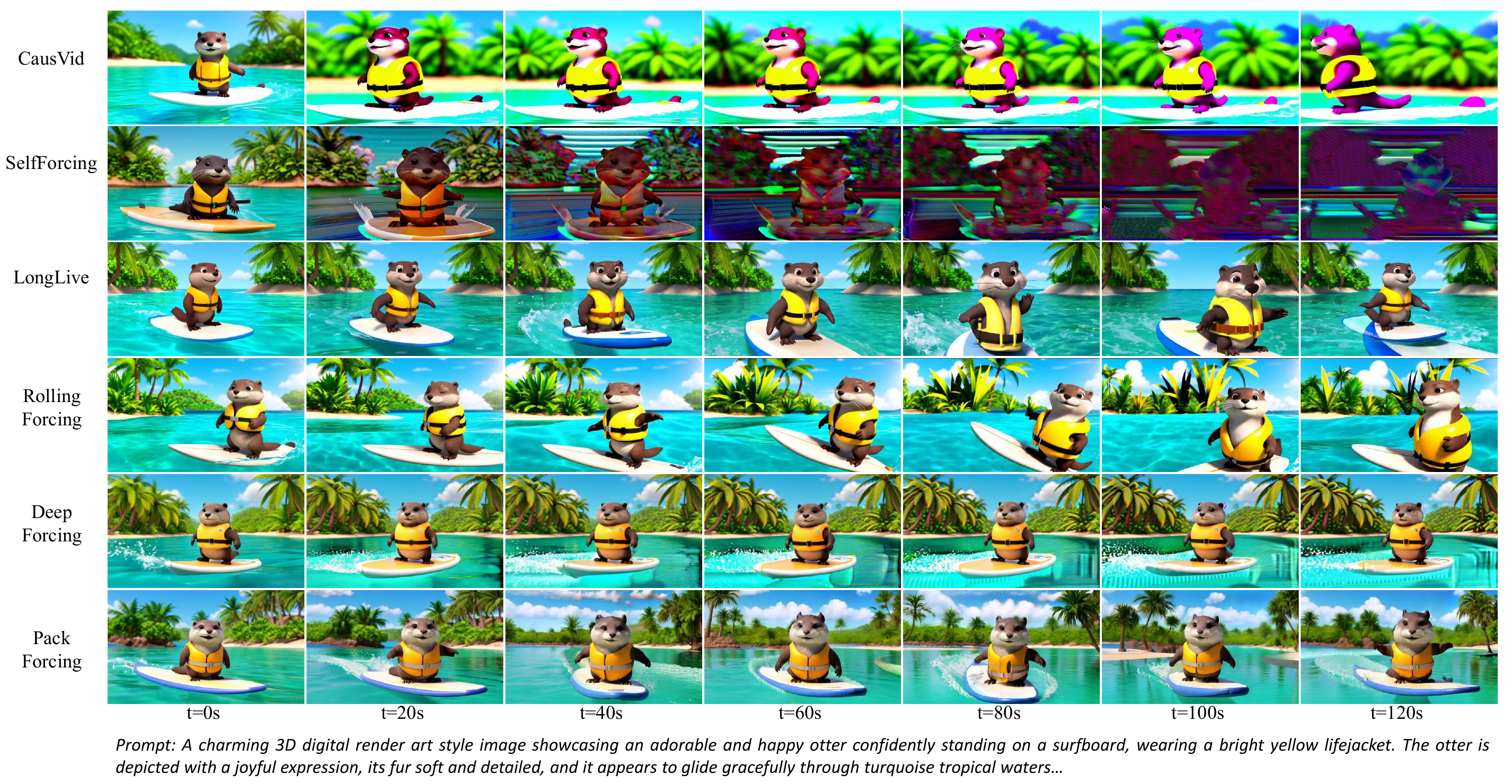}
    \caption{Our framework enables the generation of high-quality, temporally coherent videos up to 120 seconds.}
    \label{fig:teaser}
\end{figure*}

\vspace{-0em}
\renewcommand{\thefootnote}{\fnsymbol{footnote}}
\footnotetext[4]{This work was done during Xiaofeng and Shaohao's internship at Shanda AI Research Tokyo}
\footnotetext[1]{Equal contribution.}
\footnotetext[2]{Corresponding authors: mmchi@fudan.edu.cn; kaipeng.zhang@shanda.com}


\newcommand{\customsize}{\fontsize{7.7}{9}\selectfont}

\begingroup
\footnotesize
\begin{spacing}{0.88}
\end{spacing}
\endgroup

\hypersetup{linkcolor=url_color}

\newpage

\input{samplebody-journals}

\newpage

\end{document}

%% file: samplebody-journals.tex
\begin{abstract}
Autoregressive video diffusion models have demonstrated remarkable progress, yet they remain bottlenecked by intractable linear KV-cache growth, temporal repetition, and compounding errors during long-video generation. To address these challenges, we present \textbf{\methodname{}}, a unified framework that efficiently manages the generation history through a novel three-partition KV-cache strategy. Specifically, we categorize the historical context into three distinct types: (1) \textbf{Sink tokens}, which preserve early anchor frames at full resolution to maintain global semantics; (2) \textbf{Mid tokens}, which achieve a massive spatiotemporal compression (${\sim}32\times$ token reduction) via a dual-branch network fusing progressive 3D convolutions with low-resolution VAE re-encoding; and (3) \textbf{Recent tokens}, kept at full resolution to ensure local temporal coherence. To strictly bound the memory footprint without sacrificing quality, we introduce a dynamic top-$k$ context selection mechanism for the mid tokens, coupled with a continuous Temporal RoPE Adjustment that seamlessly re-aligns position gaps caused by dropped tokens with negligible overhead. Empowered by this principled hierarchical context compression, \methodname{} can generate coherent 2-minute, $832{\times}480$ videos at 16\,FPS on a single H200 GPU. It achieves a bounded KV cache of just ${\sim}\mathbf{4}$\,\textbf{GB} and enables a remarkable $\mathbf{24\times}$ temporal extrapolation ($5\,\text{s} \rightarrow 120\,\text{s}$), operating effectively either zero-shot or trained on merely 5-second clips. Extensive results on VBench demonstrate state-of-the-art temporal consistency (26.07) and dynamic degree (56.25), proving that short-video supervision is sufficient for high-quality, long-video synthesis.
\end{abstract}
\section{Introduction}
\label{sec:intro}

Recent video diffusion models~\cite{ho2022video,blattmann2023align,polyak2024movie,wan2025,valevski2024diffusion,chen2025sana,chen2024videocrafter2,ceylan2023pix2video,harvey2022flexible,wang2024motionctrl,zhang2024cameractrlii,he2024cameractrl} have demonstrated significant progress in high-fidelity and complex motion synthesis for short clips (5--15\,s). However, their bidirectional architectures typically require the simultaneous processing of all frames within a spatiotemporal volume. This computationally intensive paradigm hinders the development of streaming or real-time generation.

Autoregressive video generation~\cite{yin2025slow,huang2025self,chen2024diffusion} addresses this limitation by employing a block-by-block generation strategy. Instead of computing the entire sequence jointly, these methods sequentially cache key-value (KV) pairs from previously generated blocks to provide continuous contextual conditioning. While this approach theoretically mitigates the memory bottlenecks of joint processing and enables unbounded-length video generation, its practical application for minute-scale generation is limited by two primary challenges:

\noindent\textbf{(1)~Error accumulation.}
Small prediction errors compound iteratively during the autoregressive denoising process, leading to progressive quality degradation and semantic drift. Although Self-Forcing~\cite{huang2025self} attempts to mitigate this by training on self-generated historical frames, it still suffers from severe error accumulation beyond its training horizon. Consequently, it exhibits a significant decline in text-video alignment: within 60\,s, the model gradually loses the prompt's semantics, with its CLIP score dropping from 33.89 to 27.12 (Table~\ref{tab:clip_score_60s}).

\noindent\textbf{(2)~Unbounded memory growth.}
The KV cache scales linearly with the length of the generated video. 
For a 2-minute, $832{\times}480$ video at 16\,FPS, the full attention context grows to ${\sim}749$K tokens, requiring ${\sim}138$\,GB of KV storage across 30 transformer layers, well beyond the memory budget of a single commodity GPU.
Standard workarounds, such as history truncation~\cite{yin2025slow} or sliding windows~\cite{liu2025rolling}, severely compromise long-range coherence. Even recent advanced baselines struggle with this bottleneck. For instance, DeepForcing introduces attention sinks and participative compression to retain informative tokens based on query importance. However, to prevent unbounded KV cache expansion, it ultimately relies on aggressive buffer truncation, leading to the irreversible loss of intermediate historical memory.

A fundamental dilemma thus emerges in autoregressive video generation: mitigating error accumulation requires an extensive contextual history, yet unbounded KV cache growth inevitably forces the discarding of critical memory under hardware constraints. Maintaining a large effective context window while strictly bounding the KV cache size remains a critical open problem.

Building upon DeepForcing's insights, we recognize the effectiveness of its deep sink and participative compression mechanisms in identifying and retaining crucial historical context. However, rather than irreversibly dropping unselected intermediate tokens to save memory, we propose efficiently compressing them. To this end, we introduce \textbf{\methodname{}}, a unified framework comprehensively addressing both challenges via a principled three-partition KV cache design.

To this end, we introduce \textbf{\methodname{}}, a unified framework that comprehensively addresses both error accumulation and memory bottlenecks via a principled three-partition KV cache design. Specifically, our framework categorizes the historical context into: \textbf{(1) Sink tokens}, which preserve early anchor frames at full resolution to maintain global semantics and prevent drift; \textbf{(2) Compressed mid tokens}, which undergo a $128\times$ spatiotemporal volume compression (via a dual-branch network) to efficiently retain the bulk of the historical memory; and \textbf{(3) Recent and current tokens}, which are kept at full resolution to ensure fine-grained local coherence. 

This hierarchical design successfully bounds memory requirements while preserving critical information. To strictly limit the capacity of the compressed mid-buffer, we adapt dynamic context selection as an advanced top-$k$ selection strategy, retrieving only the most informative mid tokens during generation. To resolve the ensuing positional discontinuities caused by managing unselected tokens, we introduce a novel incremental RoPE rotation that gracefully corrects temporal positions without requiring a full cache recomputation.

In a nutshell, our primary contributions are summarized as follows:
\begin{itemize}[leftmargin=*,itemsep=2pt,topsep=3pt]
\item \textbf{Three-partition KV cache.} We propose \textbf{\methodname{}}, which partitions generation history into sink, compressed, and recent tokens, bounding per-layer attention to ${\sim}27{,}872$ tokens for any video length.

\item \textbf{Dual-branch compression.} We design a hybrid compression layer fusing progressive 3D convolutions with low-resolution re-encoding. This achieve a $128\times$ spatiotemporal compression (${\sim}32\times$ token reduction) for intermediate history, increasing effective memory capacity by over $27\times$.

\item \textbf{Incremental RoPE rotation \& Dynamic Context Selection.} We introduce a temporal-only RoPE adjustment to seamlessly correct position gaps during memory management. Alongside an importance-scored top-$k$ token selection strategy, this ensures highly stable generation over extended horizons.

\item \textbf{24$\times$ temporal extrapolation.} Trained exclusively on 5-second clips (or operating zero-shot without any training), \methodname{} successfully generates coherent 2-minute videos. It achieves state-of-the-art VBench scores and demonstrates the most stable CLIP trajectory among all compared methods.
\end{itemize}

\section{Related Work}
\label{sec:related}

\noindent \textbf{Video Diffusion Models.}
Early video models inflated 2D U-Nets with pseudo-3D modules~\cite{ho2020ddpm,rombach2022high,ho2022video,singer2022make,blattmann2023align}. Recently, Diffusion Transformers (DiTs)~\cite{peebles2023scalable,brooks2024video} have emerged as the dominant architecture, treating videos as spatiotemporal patches to enable scalable 3D attention in state-of-the-art models (e.g., CogVideoX~\cite{yang2024cogvideox}, Movie Gen~\cite{polyak2024movie}, Wan~\cite{wan2025}, Open-Sora~\cite{opensora}). Concurrently, Flow Matching~\cite{lipman2023flow,liu2022flow} has largely replaced standard diffusion to offer faster convergence. Despite these advances, current models primarily generate short clips (5--10\,s). Joint spatiotemporal modeling for minute-level videos remains computationally intractable, as full 3D attention incurs a quadratic $\mathcal{O}((THW)^2)$ memory cost. This bottleneck directly motivates the need for memory-efficient, autoregressive long-video generation strategies.

\noindent \textbf{Autoregressive Video Generation.}
Autoregressive video generation overcomes the fixed-length limitations of joint spatiotemporal modeling by synthesizing frames block-by-block and maintaining historical context via key-value (KV) caching. Recent methods have rapidly evolved this paradigm, exploring ODE-based initialization~\cite{yin2025slow}, self-generated frame conditioning~\cite{huang2025self}, rolling temporal windows~\cite{liu2025rolling}, long-short context guidance~\cite{yang2025longlive}, and enlarged attention sinks~\cite{yi2025deep}. Despite these innovations, existing approaches universally lack a mechanism to explicitly compress the KV cache. Consequently, they face a rigid trade-off: retaining the full history inevitably causes out-of-memory failures for videos exceeding roughly 80 seconds, whereas truncating the context buffer results in an irreversible loss of long-range coherence. \methodname{} explicitly breaks this memory-coherence trade-off by introducing learned spatiotemporal token compression tailored for causal video generation.

\noindent \textbf{KV Cache Management.}
KV cache management has been extensively studied in Large Language Models (LLMs) to enable long-context understanding. Representative techniques include retaining initial attention sinks~\cite{xiao2024efficient}, selecting heavy-hitter keys based on attention scores~\cite{zhang2024h2o}, and extending context via RoPE interpolation~\cite{peng2024yarn}. However, these methods primarily focus on token selection or eviction rather than explicit compression, as text representations are already highly compact. Video tokens, conversely, encode dense spatiotemporal grids characterized by massive inter-frame redundancy. Exploiting this unique structural redundancy motivates our learned $128\times$ volume compression, achieving a memory reduction far beyond what token selection alone can provide.

\noindent \textbf{Long Video Generation.}
Beyond purely autoregressive caching, traditional long video generation strategies often rely on modifying the inference noise scheduling~\cite{qiu2024freenoisetuning,ge2023preserve}, designing hierarchical planning frameworks~\cite{hong2023large}, or utilizing complex multi-stage extensions~\cite{henschel2024streamingt2v}. While effective, these methods typically require multi-stage pipelines or alter the fundamental diffusion process. \textbf{In contrast}, \methodname{} operates within a unified, single-stage causal framework. By managing the historical context through hierarchical compression and position-corrected eviction, our method achieves the generation of arbitrarily long videos with strictly bounded memory footprint and constant-time attention cost.

\section{Method}
\label{sec:method}
We first introduce the background on flow matching and causal KV caching (Sec.~\ref{sec:prelim}), then present the core components of \methodname{}: the three-partition KV cache (Sec.~\ref{sec:partition}), dual-branch compression (Sec.~\ref{sec:compression}), Dual-Resolution Shifting with incremental RoPE adjustment (Sec.~\ref{sec:fifo}), and Dynamic Context Selection (Sec.~\ref{sec:participative}).
\subsection{Preliminaries}
\label{sec:prelim}

\noindent\textbf{Flow Matching.}
Our base model builds upon the flow matching framework~\cite{lipman2023flow}.
Given a clean video latent $\mathbf{x}_0$ and standard Gaussian noise $\boldsymbol{\epsilon} \sim \mathcal{N}(\mathbf{0}, \mathbf{I})$, the noisy latent at noise level $\sigma \in [0,1]$ is constructed as:
\begin{equation}
\mathbf{x}_\sigma = (1 - \sigma)\,\mathbf{x}_0 + \sigma\,\boldsymbol{\epsilon}\,.
\label{eq:flow}
\end{equation}
A neural network $f_\theta$ is trained to predict the velocity field $\mathbf{v}_\theta(\mathbf{x}_\sigma, \sigma) \approx \boldsymbol{\epsilon} - \mathbf{x}_0$.

\noindent\textbf{KV Caching.}
A video sequence of $T$ latent frames is partitioned into non-overlapping blocks, each containing $B_f$ frames. Each block~$i$, denoted as $\mathbf{z}_i \in \mathbb{R}^{B_f \times C \times H \times W}$ (where $C$, $H$, and $W$ represent the channel, height, and width dimensions, respectively), is generated autoregressively. After spatial patchification, each block yields $n{=}B_f h w$ tokens, where $h$ and $w$ represent the spatial height and width after patchification. 

During the generation of block $i$, each transformer layer~$l$ attends to the Key-Value (KV) pairs cached from all previously generated blocks:
\begin{equation}
\mathcal{C}^l = \bigl\{(\mathbf{K}^l_j,\, \mathbf{V}^l_j)\bigr\}_{j=1}^{i-1},
\end{equation}
where $\mathbf{K}^l_j, \mathbf{V}^l_j \in \mathbb{R}^{n \times N_h \times d_h}$, with $N_h$ representing the number of attention heads and $d_h$ denoting the head dimension.

The attention operation for the current block~$i$ concatenates these historical keys and values with its own:
\begin{equation}
\operatorname{Attn}(\mathbf{Q}_i,\, \mathcal{C}^l) = \operatorname{softmax}\!\Bigl(\frac{\mathbf{Q}_i\,\mathbf{K}_{1:i}^\mid}{\sqrt{d_h}}\Bigr)\mathbf{V}_{1:i}\,,
\label{eq:full_attn}
\end{equation}
where $\mathbf{Q}_i$ is the query matrix for block $i$, while $\mathbf{K}_{1:i}$ and $\mathbf{V}_{1:i}$ represent the concatenated keys and values from block $1$ to $i$.

As generation proceeds, the KV cache grows linearly. For a 2-minute, $832{\times}480$ video at 16\,FPS ($h{=}30,\, w{=}52,\, B_f{=}4$), the context size at the final block swells to ${\sim}749$K tokens---consuming an intractable amount of GPU memory. This fundamental scaling bottleneck directly motivates our three-partition design.



\begin{figure*}[t]
\centering
\includegraphics[width=\textwidth]{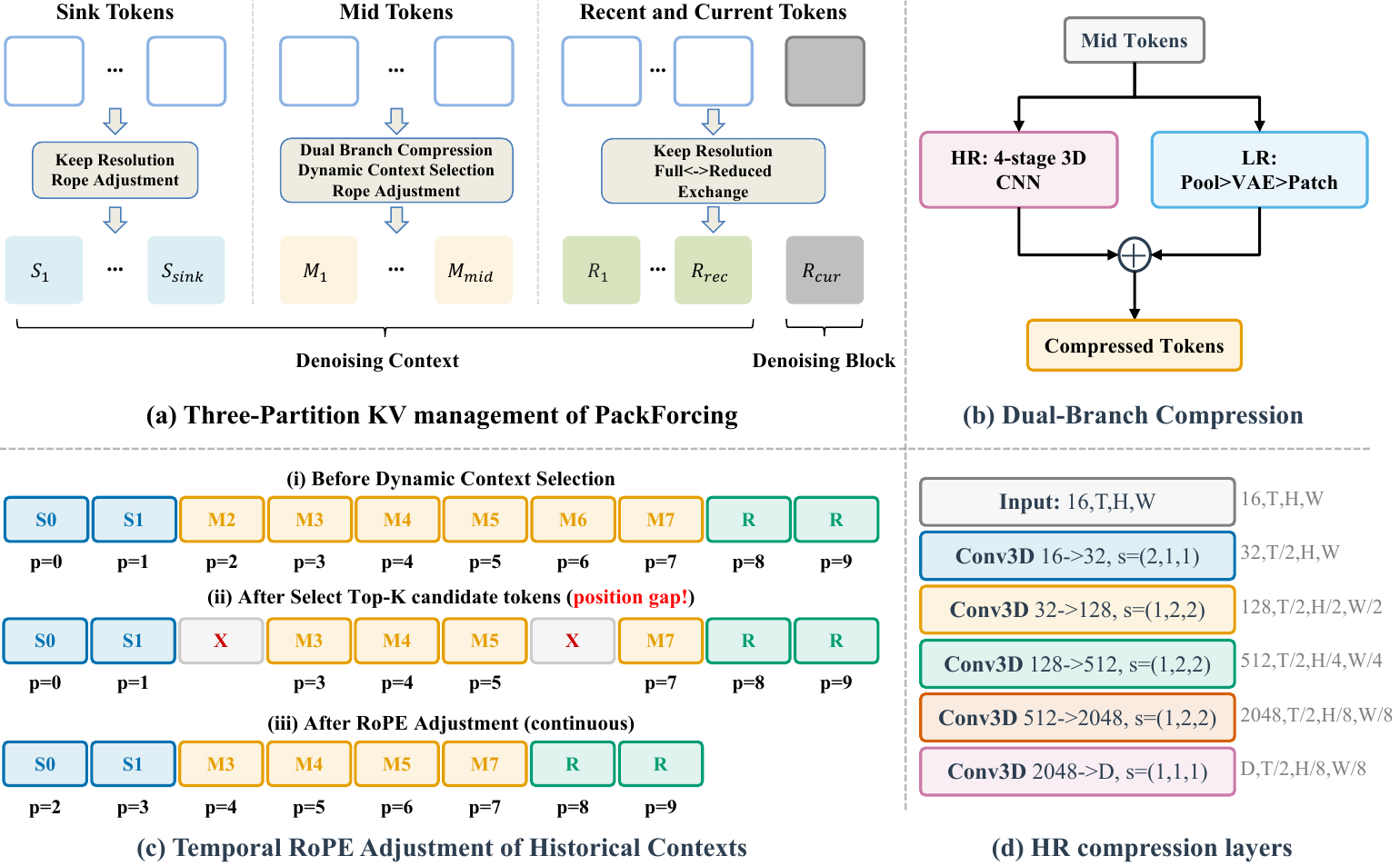}
\caption{\textbf{Overview of \methodname{}.} \textbf{(a)} The three-partition KV management organizes the denoising context into \textcolor{sinkcolor}{\textbf{sink}} tokens (full resolution), \textcolor{compcolor}{\textbf{mid}} tokens (compressed and dynamically selected), and \textcolor{recentcolor}{\textbf{recent}}/\textcolor{currentcolor}{\textbf{current}} tokens (full resolution with full-to-reduced exchange). \textbf{(b) \& (d)} The dual-branch compression module fuses a High-Resolution (HR) branch (a progressive 4-stage 3D CNN) with a Low-Resolution (LR) branch (pooling followed by VAE re-encoding) via element-wise addition to minimize token count. \textbf{(c)} To efficiently bound the memory footprint, dynamic context selection is applied to retain only the top-$K$ informative mid tokens. To resolve the ensuing position gaps caused by dropped tokens, a Temporal RoPE Adjustment is utilized to re-align and enforce continuous positional indices across the historical contexts.}
\label{fig:teaser}
\end{figure*}
\subsection{Three-Partition KV Cache}
\label{sec:partition}

The core idea of \methodname{} is to decouple the monotonically growing generation history into three distinct \emph{functional} partitions. Rather than applying a one-size-fits-all eviction or compression strategy, we apply a tailored policy to each partition based on its temporal role and information density (Fig.~\ref{fig:teaser}).

\noindent \textbf{Sink Tokens (Full resolution, never evicted).}
Inspired by the attention-sink phenomenon in StreamingLLM~\cite{xiao2024efficient}, we hypothesize that the earliest generated frames serve as critical semantic anchors. Let $N_\text{sink}$ denote the number of these initial frames, corresponding to the first $N_\text{sink}/B_f$ generation blocks. For a given transformer layer $l$, the sink cache is defined as:
\begin{equation}
\mathcal{C}_\text{sink}^l = \bigl\{(\mathbf{K}^l_j, \mathbf{V}^l_j)\bigr\}_{j=1}^{N_\text{sink}/B_f}\,,\quad
|\mathcal{C}_\text{sink}^l| = \tfrac{N_\text{sink}}{B_f}\,n\,,
\end{equation}
where $j$ is the block index, and $\mathbf{K}^l_j, \mathbf{V}^l_j$ are the original, uncompressed key and value. These tokens lock in the scene layout, subject identity, and global style. Because they are vital for preventing semantic drift, they are \emph{never compressed or evicted}. We set $N_\text{sink}{=}8$ (two blocks), which consumes ${<}2\%$ of the total token budget for a 2-minute video, yet provides a robust and stable global reference throughout the entire generation process.

\noindent \textbf{Compressed Mid Tokens (${\sim}32\times$ token reduction \& dynamically routed).}
The vast majority of the video history falls between the initial sink frames and the most recent window. We define this region as the \emph{mid} partition. Retaining this partition at full resolution is computationally prohibitive and highly redundant. Instead, tokens populating this region are represented by highly compressed KV pairs $(\tilde{\mathbf{K}}^l_j, \tilde{\mathbf{V}}^l_j)$ produced via our dual-branch module (Sec.~\ref{sec:compression}). Furthermore, as this compressed buffer accumulates over time, we do not attend to the entire pool indiscriminately. We employ \emph{Dynamic Context Selection} (Sec.~\ref{sec:participative}) to dynamically evaluate query-key affinities, actively routing only the $N_\text{mid}$ most informative blocks to form the active set $\mathcal{S}_\text{mid}$ for the current computation:
\begin{equation}
\mathcal{C}_\text{mid}^l = \bigl\{(\tilde{\mathbf{K}}^l_j, \tilde{\mathbf{V}}^l_j)\bigr\}_{j \in \mathcal{S}_\text{mid}}\,,\quad
|\mathcal{C}_\text{mid}^l| \le N_\text{mid}\cdot N_c\,,
\end{equation}
where the tilde ($\sim$) denotes compressed part and $N_\text{mid}$ limits the active computational budget. $N_c$ is the token count per compressed block, calculated as $N_c = \lfloor B_f/2\rfloor \times \lfloor h/4\rfloor \times \lfloor w/4\rfloor$. Here, the factors $1/2$ and $1/4$ correspond to the downsampling strides of the compression module along the temporal and spatial dimensions. With default settings ($B_f{=}4,\, h{=}30,\, w{=}52$), each block is compressed to $N_c = 182$ tokens---a dramatic ${\sim}32\times$ reduction from the original $n{=}6{,}240$ tokens.

\noindent \textbf{Recent \& Current Tokens (Dual-resolution shifting).}
To maintain high-fidelity local temporal dynamics when generating new video frames, the most recently generated frames must be kept pristine. Let $i$ denote the index of the block currently being generated and $N_\text{recent}$ be the number of preceding recent frames. The context for this partition comprises the intact KV pairs from these recent blocks, alongside the current block $i$ itself:
\begin{equation}
\mathcal{C}_\text{rc}^l = \bigl\{(\mathbf{K}^l_j, \mathbf{V}^l_j)\bigr\}_{j=i-N_\text{recent}/B_f}^{i}\,,\quad
|\mathcal{C}_\text{rc}^l| = \bigl(\tfrac{N_\text{recent}}{B_f}+1\bigr)\,n\,.
\end{equation}
Preserving these recent tokens at uncompressed resolution guarantees smooth temporal transitions. Crucially, to bridge this partition with the mid-buffer without incurring sequential latency, we concurrently compute a low-resolution backup for these tokens. As detailed in Sec.~\ref{sec:fifo}, this dual-resolution shifting pipeline perfectly hides the compression overhead and ensures a seamless transition of aging recent tokens into long-term mid memory.

\noindent \textbf{Bounded Attention Context.}
During the generation of block $i$, the transformer layer $l$ concatenates the three partitions to form the active attention context:
\begin{equation}
\mathcal{C}^l = \mathcal{C}^l_\text{sink} \;\|\; \mathcal{C}^l_\text{mid} \;\|\; \mathcal{C}^l_\text{rc}\,,
\label{eq:full_context}
\end{equation}
which enforces a \emph{constant} token count for the attention computation:
$|\mathcal{C}^l| = \frac{N_\text{sink}}{B_f}\,n + N_\text{mid}\cdot N_c + \bigl(\frac{N_\text{recent}}{B_f}+1\bigr)\,n$.
Crucially, while the entire generation history is persistently maintained within the memory buffers (either at full resolution or in a highly compressed state), the actual context input for generating the block $i$ is strictly bounded and independent of the total video length $T$. Rather than attending to the full growing sequence, this fixed-size input context is dynamically retrieved from the comprehensive historical partitions, ensuring $O(1)$ attention complexity without discarding any past memory.
\subsection{Dual-Branch HR Compression}
\label{sec:compression}

The mid partition requires a massive token reduction (${\sim}32\times$) while retaining sufficient structural and semantic information for coherent attention patterns (see Fig.~\ref{fig:attention_analysis}). A single-pathway compressor faces a steep trade-off: aggressive spatial downsampling preserves layout but loses texture, whereas semantic pooling preserves meaning but destroys spatial structure. To resolve this, we propose a \emph{dual-branch} compression module (Fig.~\ref{fig:teaser}(b)) that aggregates fine-grained structure (HR branch) and coarse semantics (LR branch).

\noindent \textbf{HR Branch: Progressive 3D Convolution.}
The HR branch operates directly on the VAE latent $\mathbf{z} \in \mathbb{R}^{B \times C \times T \times H \times W}$ to preserve local, fine-grained details. It applies a cascade of strided 3D convolutions with SiLU activations. Specifically, it first performs a $2\times$ temporal compression, followed by three stages of $2\times$ spatial compression, and a final $1{\times}1{\times}1$ projection to the model's hidden dimension $d{=}1536$. This yields a structurally rich representation $\mathbf{h}_\text{HR}$ with a total $128\times$ volume reduction ($2{\times}8{\times}8$) in the latent space.

\noindent \textbf{LR Branch: Pixel-Space Re-encoding.}
To capture complementary global context, the LR branch operates via a distinct pixel-space pathway. We decode the latent $\mathbf{z}$ back into pixel frames, apply a 3D average pooling ($2\times$ temporally, $4\times$ spatially), and then re-encode the pooled frames back into the latent space using the frozen VAE encoder, followed by standard patch embedding to obtain $\mathbf{h}_\text{LR}$. This decoding-pooling-encoding pipeline preserves the perceptual layout far better than direct pooling in the latent space.

\noindent \textbf{Feature Fusion.}
The outputs from both branches share the same dimensional space and are fused via element-wise addition:
\begin{equation}
\tilde{\mathbf{h}} = \mathbf{h}_\text{HR} + \mathbf{h}_\text{LR} \in \mathbb{R}^{B \times N_c \times d}\,,
\label{eq:fusion}
\end{equation}
where the compressed token count is $N_c = \lfloor T/2\rfloor \times \lfloor H/8\rfloor \times \lfloor W/8\rfloor$. Given that the original patch embedding already performs a $1{\times}2{\times}2$ spatial reduction, our dual-branch module effectively achieves a net token reduction of ${\sim}32\times$ per block (e.g., from $6{,}240$ to $182$ tokens). This simple yet effective fusion ensures comprehensive information retention under extreme compression.

\subsection{Dual-Resolution Shifting and Incremental RoPE adjustment}
\label{sec:fifo}

\noindent \textbf{Dual-Resolution Shifting Mechanism.}
Unlike FIFO methods that permanently discard tokens, we preserve long-term memory via a seamless dual-resolution pipeline. During chunk generation, we concurrently compute a full-resolution KV cache for immediate prediction and a reduced-resolution backup. Once the next chunk is generated, new full-resolution tokens replace the old ones, while the pre-computed compressed tokens smoothly slide into the \emph{mid} partition. This retains comprehensive history while hiding compression latency.

\noindent \textbf{The Position Misalignment Problem.}
Although this dual-track pipeline efficiently populates the compressed mid-buffer, strictly bounding the total memory capacity ($N_\text{mid}$) eventually necessitates evicting the absolute oldest compressed blocks from the mid partition. 
Our backbone uses 3D Rotary Position Embeddings (RoPE)~\cite{su2024roformer} with separate temporal, height, and width frequencies $\boldsymbol{\theta} = [\boldsymbol{\theta}_t,\, \boldsymbol{\theta}_h,\, \boldsymbol{\theta}_w]$, allocated as $44 + 42 + 42 = 128$ dimensions per head.
When a key is cached, it already carries the rotation for its absolute position $p$:
\begin{equation}
\mathbf{k}_{\text{cached}}^{(p)} = \mathbf{k}_{\text{raw}} \odot e^{i\,\boldsymbol{\theta}_p}\,.
\label{eq:rope_cached}
\end{equation}
After evicting $\Delta$ blocks ($\delta = \Delta B_f$ frames) to maintain the capacity budget, the \emph{sink} keys still encode positions $0,\dots, N_\text{sink}{-}1$, yet the earliest surviving \emph{mid} key now encodes position $N_\text{sink} + \delta$.
The resulting position gap breaks the positional continuity that the transformer attention relies on.

\noindent \textbf{Incremental RoPE adjustment.}
We exploit two properties to resolve this without a full recomputation: (i)~RoPE adjustments are \emph{multiplicative}: $e^{i\boldsymbol{\theta}_p}\cdot e^{i\boldsymbol{\theta}_\delta} = e^{i\boldsymbol{\theta}_{p+\delta}}$; and (ii)~eviction shifts only the \emph{temporal} axis.
We therefore apply a highly efficient, temporal-only RoPE adjustment to the sink keys:
\begin{equation}
\mathbf{k}_{\text{sink}}' = \mathbf{k}_{\text{sink}} \odot \,e^{i\,\boldsymbol{\theta}_t(\delta)},\; \mathbf{1}_h,\; \mathbf{1}_w\,\,,
\label{eq:rope_correct}
\end{equation}
where $\mathbf{1}_h,\mathbf{1}_w$ are identity (unit-magnitude) rotations that leave spatial positions unchanged.
After adjustment, the sink positions become $\delta, \delta{+}1, \dots$, seamlessly preceding the mid positions.
This operation is applied once per eviction event across all layers, costing ${<}0.1\%$ of total inference time (Table~\ref{tab:memory} in Appendix).





\subsection{Dynamic Context Selection}
\label{sec:participative}

Treating all compressed blocks equally ignores their varying semantic importance. To prioritize visually critical keyframes, we introduce a dynamic context selection mechanism based on query-key affinity. Unlike DeepForcing~\cite{yi2025deep}, which permanently prunes unselected tokens, we employ a non-destructive soft-selection. We retrieve only the top-$K$ most relevant mid-blocks for attention, keeping unselected tokens safely archived for potential future reactivation as the scene evolves. To ensure negligible overhead (${<}1\%$), affinity scoring occurs only at the first denoising step of each block, caching the indices for subsequent steps. We further accelerate this by subsampling query tokens and using half the attention heads. This soft-routing improves subject consistency by +0.8 and the CLIP score by +0.12 over strict FIFO eviction (Table~\ref{tab:ablation_eviction}).

\begin{figure*}[h]
    \centering
    \includegraphics[width=1.0\textwidth]{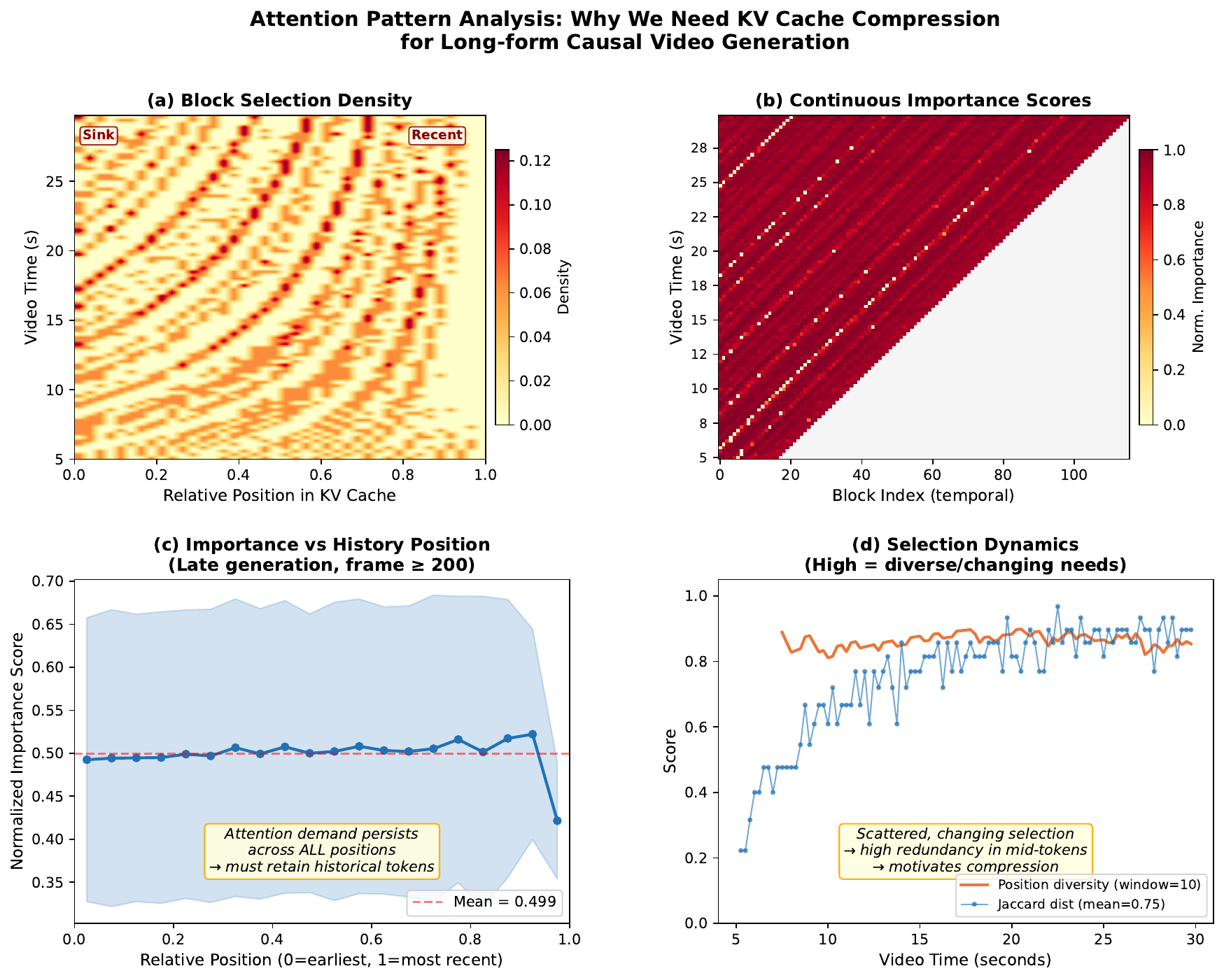}
    \caption{\textbf{Attention patterns in causal video generation.} 
    \textbf{(a)}~Density map shows attention distributed across the entire history. 
    \textbf{(b)}~Importance scores $\phi_j = \sum_i \mathbf{q}_i^\top \mathbf{k}_j / \sqrt{d}$ reveal sparse information. 
    \textbf{(c)}~Near-flat late-stage importance (mean\,$=$\,$0.499$) rules out FIFO eviction. 
    \textbf{(d)}~High Jaccard distance ($0.75$) and position diversity ($>0.85$) show rapid, diverse block retrieval, motivating compression.
    }
    \label{fig:attention_analysis}
\end{figure*}
\subsection{Empirical Analysis of Temporal Attention Patterns}
\label{sec:attention_analysis}

To motivate our KV cache design, we empirically investigate the attention distribution of the denoising network during a 480-frame generation (Fig.~\ref{fig:attention_analysis}). Our analysis reveals two critical insights (detailed in the Appendix): (1) Attention demand persists across the entire video history, invalidating naive FIFO eviction strategies; and (2) Highly attended tokens are sparsely and dynamically distributed, exhibiting a high Jaccard distance (0.75) between consecutive selection steps. These observations fundamentally justify our three-partition cache architecture (sink, mid, and recent). By aggressively compressing the sporadically queried yet globally essential mid-range tokens, we successfully preserve comprehensive spatiotemporal context within a strictly bounded memory footprint.

.

\section{Experiments}
\label{sec:exp}
\subsection{Experimental Settings}

\noindent \textbf{Implementation Details.}
\methodname{} employs the Wan2.1-T2V-1.3B~\cite{wan2025} backbone to generate $832{\times}480$ videos at 16\,FPS. Text conditioning relies on the UMT5-XXL encoder, where text key-value pairs are computed once and cached globally to reduce overhead. Consistent with recent causal generative frameworks~\cite{yin2025slow, huang2025self, liu2025rolling, yi2025deep}, the causal student is initialized via ODE trajectory alignment. Training prompts are sourced from VidProM and augmented via large language models following the Self-Forcing paradigm. For the cache partitions, we set $N_\text{sink}{=}8$, $N_\text{recent}{=}4$, and $N_\text{top}{=}16$. Chunk-wise generation operates with $B_f{=}4$ latent frames per block and $S{=}4$ distilled denoising steps. The model is trained for 3{,}000 iterations with a batch size of 8 on a 20-latent-frame temporal window (\~5 seconds). We use AdamW~\cite{loshchilov2019adamw} ($\beta_1{=}0, \beta_2{=}0.999$), setting learning rates to $2.0{\times}10^{-6}$ for the generator $G_\theta$ and $1.0{\times}10^{-6}$ for the fake score estimator $s_\text{fake}$, with a 1:5 update ratio. During inference, we utilize a classifier-free guidance scale of $3.0$ and a timestep shift of $5.0$.

\noindent \textbf{Evaluation Protocols.}
To rigorously assess long video generation performance, we adopt the evaluation setting of VBench-Long~\cite{huang2025vbench++}. Specifically, we utilize 128 text prompts sourced from MovieGen~\cite{polyak2024movie}, adhering strictly to the prompt selection protocol established in Self-Forcing++~\cite{cui2025self}. Consistent with the standard Self-Forcing~\cite{cui2025self} paradigm, each prompt is refined and expanded using Qwen2.5-7B-Instruct~\cite{qwen2025qwen25technicalreport} prior to generation. Under this standardized setup, we generate videos at both 60\,s and 120\,s durations. We quantitatively evaluate the results using seven core metrics from the VBench framework~\cite{huang2024vbench}: Dynamic Degree (Dyn.~Deg.), Motion Smoothness (Mot.~Smth.), Overall Consistency (Over.~Cons.), Imaging Quality (Img.~Qual.), Aesthetic Quality (Aest.~Qual.), Subject Consistency (Subj.~Cons.), and Background Consistency (Back.~Cons.). Furthermore, to measure the temporal stability of text-video alignment over extended durations, we compute and report CLIP scores at 10-second intervals throughout the generated sequences.

\begin{table*}[!h]
\centering
\caption{\textbf{Quantitative comparison} on 60\,s and 120\,s benchmarks (7 VBench~\cite{huang2024vbench} metrics). Best results are highlighted in \textbf{bold}.}
\label{tab:performance_comparison}
\scriptsize
\renewcommand{\arraystretch}{1.25}

\resizebox{\textwidth}{!}{
\begin{tabular}{@{}l|ccccccc@{}}
\thickhline
Method & \makecell{Dyn.\\Deg.}$\uparrow$ & \makecell{Mot.\\Smth.}$\uparrow$ & \makecell{Over.\\Cons.}$\uparrow$ & \makecell{Img.\\Qual.}$\uparrow$ & \makecell{Aest.\\Qual.}$\uparrow$ & \makecell{Subj.\\Cons.}$\uparrow$ & \makecell{Back.\\Cons.}$\uparrow$ \\
\hline
\multicolumn{8}{c}{\textit{60\,s generation}} \\
\hline
CausVid~\cite{yin2025slow}         & 48.43 & 98.04 & 23.36 & 65.69 & 60.63 & 84.53 & 89.84 \\
LongLive~\cite{yang2025longlive}   & 44.53 & 98.70 & 25.73 & 69.06 & \textbf{63.30} & {92.00} & 92.97 \\
Self-Forcing~\cite{huang2025self}  & 35.93 & 98.26 & 24.92 & 66.62 & 57.15 & 80.41 & 86.95 \\
Rolling Forcing~\cite{liu2025rolling}   & 33.59 & \textbf{98.70} & 25.73 & \textbf{71.06} & 61.43 & 91.62 & {93.00} \\
Deep Forcing~\cite{yi2025deep}   & 53.67  & 98.56 & 21.75 & 67.75 & 58.88 & \textbf{92.55} & \textbf{93.80} \\
\textbf{\methodname{} (ours)}      & \textbf{56.25} & 98.29 & \textbf{26.07} & 69.36 & 62.56 & 90.49 & 93.46 \\
\hline
\multicolumn{8}{c}{\textit{120\,s generation}} \\
\hline
CausVid~\cite{yin2025slow}         & 50.00 & 98.11 & 23.13 & 65.41 & 60.11 & 83.24 & 87.83 \\
LongLive~\cite{yang2025longlive}   & 44.53 & \textbf{98.72} & 25.95 & 69.59 & \textbf{63.00} & {91.54} & \textbf{93.73} \\
Self-Forcing~\cite{huang2025self}  & 30.46 & 98.12 & 23.42 & 62.49 & 51.68 & 74.40 & 83.57 \\
Rolling Forcing~\cite{liu2025rolling}   & 35.15 & 98.65 & 25.45 & \textbf{70.58} & 60.62 & 90.14 & 92.40 \\
Deep Forcing~\cite{yi2025deep}   & 52.84 & 98.22 & 21.38 & 68.21 & 57.96 & 91.95 & 92.55 \\
\textbf{\methodname{} (ours)}      & \textbf{54.12} & 98.35 & \textbf{26.05} & 69.67 & 61.98 & \textbf{92.84} & 91.88 \\
\thickhline
\end{tabular}
} 

\end{table*}
\subsection{Main Results}
\noindent \textbf{Quantitative Comparison.}
Table~\ref{tab:performance_comparison} evaluates 60\,s and 120\,s video generation across seven VBench metrics. \methodname{} excels in motion synthesis, achieving the highest Dynamic Degree at both durations (56.25 and 54.12) and outperforming the strongest baseline, CausVid, by +7.82 and +4.12. This confirms that our persistent memory mechanism provides reliable temporal grounding, enabling the model to confidently synthesize extensive motion. Furthermore, \methodname{} demonstrates superior stability over extended horizons. While baselines like Self-Forcing degrade significantly from 60\,s to 120\,s (e.g., Subject Consistency drops by 6.01, Aesthetic Quality by 5.47), our performance declines are marginal (-1.01 and -0.49, respectively). This robustness highlights the effectiveness of sink tokens in anchoring global semantics and the dynamically routed mid-buffer in preserving intermediate context. Finally, despite an aggressive ${\sim}32\times$ token reduction, \methodname{} maintains highly competitive Image and Aesthetic Quality, proving that our dual-branch architecture successfully retains perceptually critical spatiotemporal details.

\noindent \textbf{Long-Range Consistency.} 
Table~\ref{tab:clip_score_60s} evaluates the temporal stability of text-video alignment via CLIP scores at 10-second intervals. \methodname{} maintains the highest alignment throughout the 60-second generation, with a marginal decline of only 1.14 points (from 34.04 to 32.90). Conversely, baselines suffer from compounding errors: Self-Forcing exhibits a severe 6.77-point drop, while CausVid declines by 1.86 points. This sustained temporal coherence directly validates our sink token mechanism's ability to anchor global semantics across extended horizons.

\begin{table*}[!h]
\centering
\caption{CLIP Score comparison on long video generation (60s).}
\label{tab:clip_score_60s}

\begin{tabular}{@{}lccccccc@{}}
\thickhline
Method & 0--10\,s & 10--20\,s & 20--30\,s & 30--40\,s & 40--50\,s & 50--60\,s & Overall \\
\hline
CausVid~\cite{yin2025slow}        & 32.65 & 31.78 & 31.47 & 31.13 & 30.81 & 30.79 & 31.44 \\
LongLive~\cite{yang2025longlive}  & 33.95 & 33.38 & 33.14 & \textbf{33.51} & \textbf{33.45} & \textbf{33.36} & 33.46 \\
Self-Forcing~\cite{huang2025self} & 33.89 & 33.23 & 31.66 & 29.99 & 28.37 & 27.12 & 30.71 \\
Rolling Forcing~\cite{liu2025rolling}  & 33.85 & 33.39 & 32.94 & 32.78 & 32.49 & 32.25 & 32.95 \\
Deep Forcing~\cite{yi2025deep}  & 33.47 & 33.29 & 32.38 & 32.28 & 32.26 & 32.27 & 32.33 \\
\textbf{\methodname{} (ours)}     & \textbf{34.04} & \textbf{33.99} & \textbf{33.70} & {33.37} & {33.24} & {32.90} & \textbf{33.54} \\
\thickhline
\end{tabular}

\end{table*}

\begin{figure*}[t]
\centering
\includegraphics[width=\linewidth]{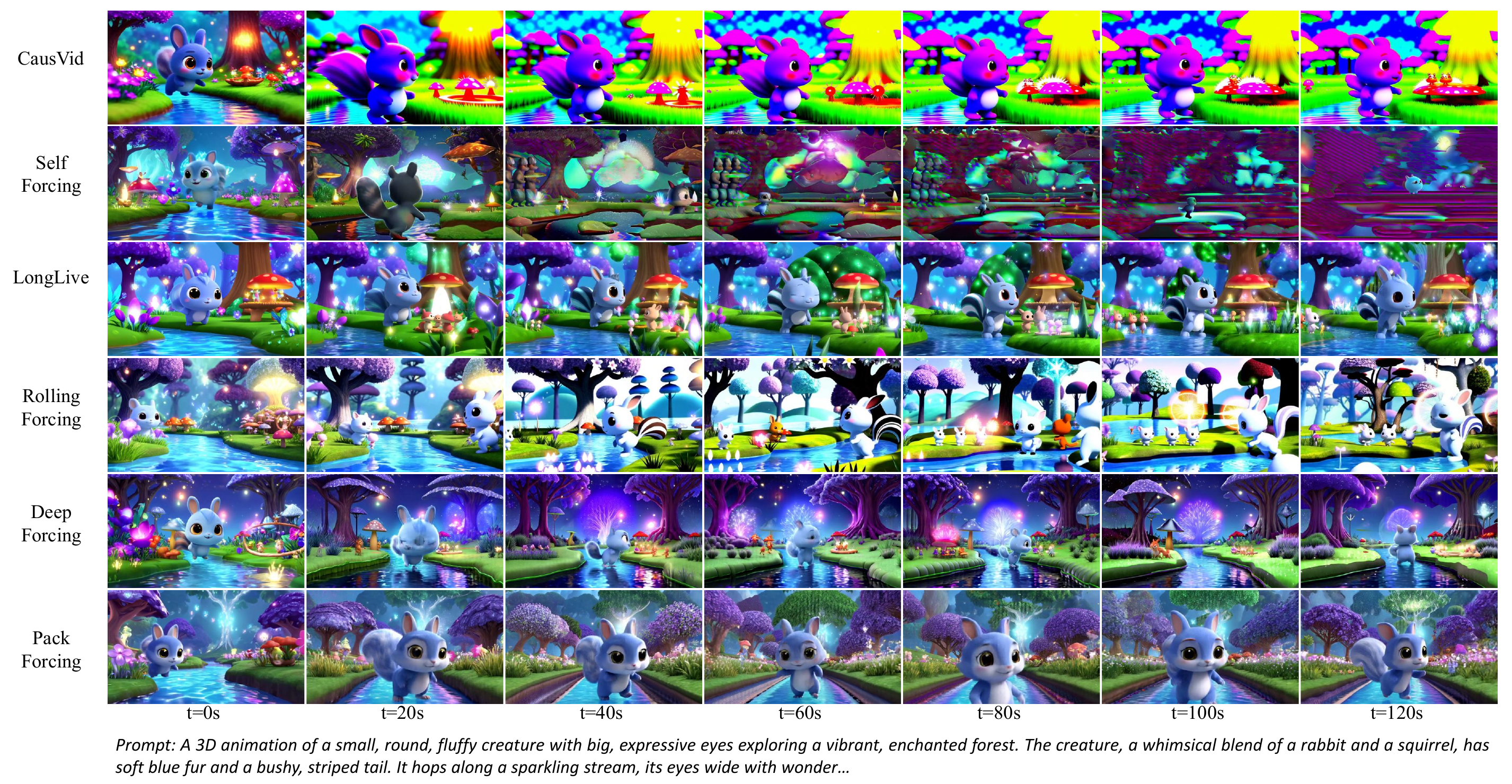}
\caption{\textbf{Qualitative comparison of 120\,s generation.} Sampled frames across seven timestamps under the same prompt. \methodname{} consistently maintains strict subject identity and high visual fidelity throughout the entire sequence. In contrast, baseline methods suffer from progressive degradation over time: Self-Forcing exhibits color shifts by 60\,s and eventual collapse, CausVid loses background details, and LongLive generates severely restricted motion. Furthermore, Rolling-Forcing struggles with significant subject inconsistencies, while DeepForcing loses the main subject at 100s.}
\label{fig:qualitative}
\end{figure*}

\noindent \textbf{Qualitative Comparison.}
Fig.~\ref{fig:qualitative} presents sampled frames from a 120-second generation for all methods under the same text prompt. At 0\,s all methods produce comparable quality. By 60\,s, Self-Forcing exhibits visible color shift and object duplication; CausVid loses fine details in the background. At 120\,s, only \methodname{} and LongLive maintain coherent subjects, but LongLive shows noticeably less camera and subject motion. \methodname{} preserves both subject identity and dynamic motion throughout the full two minutes, thanks to the persistent sink tokens and compressed history.

\subsection{Ablation Studies}

\begin{figure*}[t]
\centering
\includegraphics[width=\textwidth]{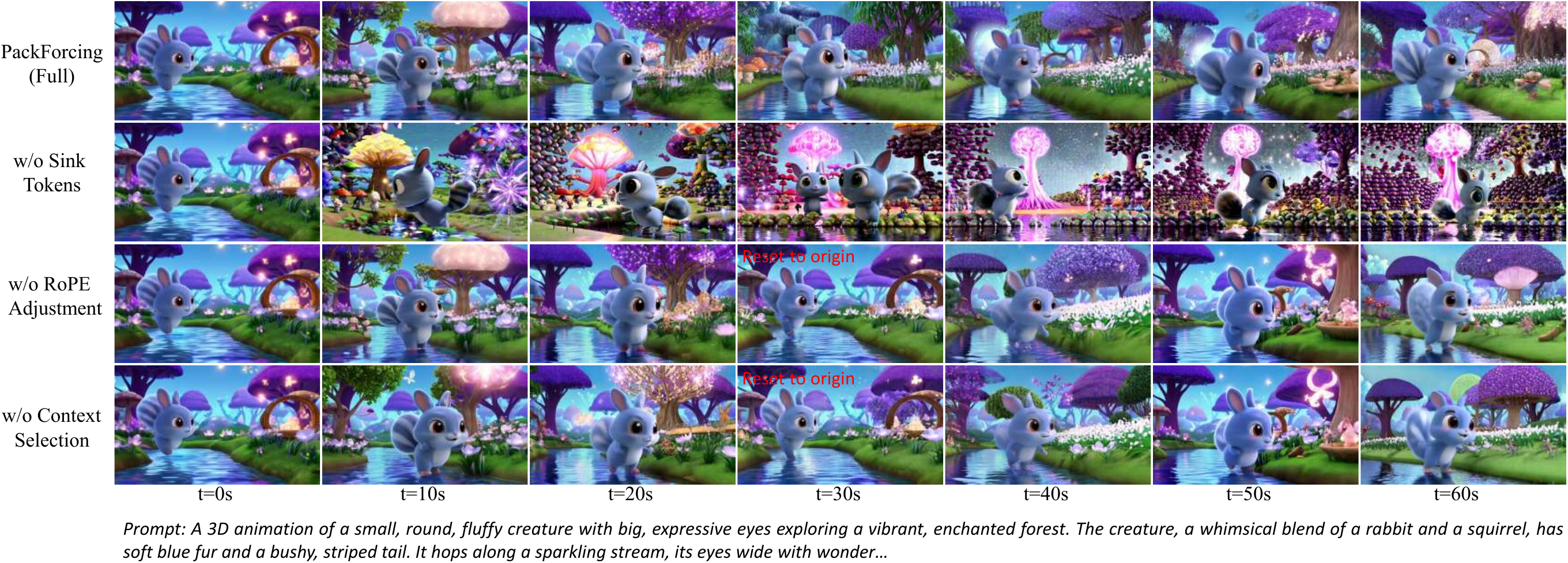}
\caption{\textbf{Qualitative ablation results on 60\,s generation.} Removing sink tokens leads to progressive semantic drift, whereas disabling either the RoPE adjustment or dynamic context selection introduces severe frame reset artifacts.}
\label{fig:ablation_visual}
\end{figure*}

We perform systematic ablations on the 60\,s benchmark to evaluate each critical component of \methodname{}. Qualitative comparisons are shown in Fig.~\ref{fig:ablation_visual}.

\begin{table}[h]
\centering
\caption{\textbf{Quantitative ablation results of sink size.} We consolidate overall CLIP scores and VBench metrics across varying sink sizes. Setting $N_\text{sink}{=}8$ achieves the optimal balance between dynamic motion and semantic consistency.}
\label{tab:sink_ablation}

\begin{tabular}{@{}c|c|ccccccc@{}}
\thickhline
Sink Size & \makecell{Overall\\CLIP}\,$\big\uparrow$ & \makecell{Dyn.\\Deg.}\,$\big\uparrow$ & \makecell{Mot.\\Smth.}\,$\big\uparrow$ & \makecell{Over.\\Cons.}\,$\big\uparrow$ & \makecell{Img.\\Qual.}\,$\big\uparrow$ & \makecell{Aest.\\Qual.}\,$\big\uparrow$ & \makecell{Subj.\\Cons.}\,$\big\uparrow$ & \makecell{Back.\\Cons.}\,$\big\uparrow$ \\
\hline
0  & 31.24 & 20.31 & \textbf{98.89} & 23.11 & 72.87 & 59.37 & 74.72 & 86.37 \\
2  & 34.85 & 49.69 & 98.57 & 26.78 & 72.59 & 65.85 & 91.37 & 93.68 \\
4  & 34.85 & \textbf{50.16} & 98.57 & \textbf{26.99} & 72.59 & 65.84 & 91.37 & 93.73 \\
8  & 35.09 & 49.84 & 98.68 & 26.29 & \textbf{73.18} & \textbf{66.46} & 93.11 & 94.53 \\
16 & \textbf{35.39} & 35.16 & 98.82 & 26.71 & 73.05 & 66.34 & \textbf{93.84} & \textbf{94.92} \\
\thickhline
\end{tabular}

\end{table}

\noindent \textbf{Effect of Sink Tokens.}
To evaluate the impact of sink size ($N_\text{sink}$) on long-term stability, we conduct an ablation study on 60 randomly selected samples (Table~\ref{tab:sink_ablation}). Removing attention sinks entirely ($N_\text{sink}{=}0$) causes severe semantic drift, evidenced by sharp drops in the Overall CLIP score (35.09 to 31.24) and Subject Consistency (93.11 to 74.72), confirming their role as critical global anchors (Fig.~\ref{fig:ablation_visual}). Conversely, an excessively large sink ($N_\text{sink}{=}16$) maximizes consistency but stifles motion, plummeting the Dynamic Degree to 35.16 as the model over-conditions on static early frames. Setting $N_\text{sink}{=}8$ achieves the optimal balance. It preserves dynamic richness (49.84) and yields the best Image (73.18) and Aesthetic (66.46) Quality, with Subject Consistency within $1\%$ of the $N_\text{sink}{=}16$ setting—all while consuming strictly bounded memory (${<}2\%$ of the total token budget). Temporal CLIP details can be found in the Appendix.

\noindent \textbf{Effect of Compression Branches.}
Table~\ref{tab:ablation_branch} isolates each compression branch, evaluated at 60\,s where all variants fit in memory. The HR branch alone provides strong spatial compression but misses coarse semantic cues; the LR branch alone preserves semantics but lacks spatial precision. 


\begin{table}[t]
\centering

\begin{minipage}[t]{0.48\textwidth}
\centering
\caption{Ablation on compress. branches.}
\label{tab:ablation_branch}

\begin{tabular}{@{}lcccc@{}}
\thickhline
Branch & \makecell{Img.\\Qual.}\,$\big\uparrow$& \makecell{Over.\\Cons.}\,$\big\uparrow$ & CLIP\,$\big\uparrow$ & 60\,s \\
\hline
HR only       & 68.12 & 25.41 & 32.97 & \cmark \\
LR only       & 67.45 & 25.18 & 33.11 & \cmark \\
\textbf{HR + LR} & \textbf{69.36} & \textbf{26.07} & \textbf{33.54} & \cmark \\
\thickhline
\end{tabular}

\end{minipage}%
\hfill 
\begin{minipage}[t]{0.48\textwidth}
\centering
\caption{Ablation on eviction strategy.}
\label{tab:ablation_eviction}
\begin{tabular}{@{}lccc@{}}
\thickhline
Strategy & \makecell{Subj.\\Cons.}\,$\big\uparrow$ & \makecell{Over.\\Cons.}\,$\big\uparrow$ & CLIP\,$\big\uparrow$ \\
\hline
Random        & 86.31 & 25.42 & 33.01 \\
FIFO          & 87.82 & 25.91 & 33.42 \\
\textbf{Dynamic Select} & \textbf{88.62} & \textbf{26.07} & \textbf{33.54} \\
\thickhline
\end{tabular}

\end{minipage}

\end{table}

\noindent \textbf{FIFO vs.\ Dynamic Context Selection.}
Table~\ref{tab:ablation_eviction} compares memory management strategies within the compressed mid-buffer. Dynamic context selection outperforms standard FIFO, yielding notable improvements in Subject Consistency (+0.8) and the overall CLIP score (+0.12). This advantage stems from its affinity-driven approach, which prioritizes the retention of highly attended historical blocks rather than relying on a rigid chronological eviction.

\subsection{Discussion}
\noindent \textbf{Generalization from Short-Video Supervision.}
The remarkable $24\times$ temporal extrapolation capability of \methodname{} (from 5\,s training clips to 120\,s generated videos) can be attributed to two primary mechanisms. First, the framework enforces {context size invariance}. By systematically compressing and managing the KV cache, the attention context remains strictly bounded (e.g., ${\sim}27{,}872$ tokens) during both training and inference. This effectively prevents out-of-distribution sequence length shifts that typically trigger error accumulation in standard autoregressive models. Second, the architecture ensures {representational compatibility}. Jointly training the dual-branch compression layer aligns the compressed tokens with the full-resolution tokens within the same latent subspace, enabling the transformer to seamlessly attend to the highly compressed historical features.

\noindent \textbf{Motion Richness and Dynamic Degree.}
As indicated by the VBench evaluations, \methodname{} consistently achieves superior dynamic degree scores. Existing autoregressive baselines, which often lack persistent long-range memory, tend to degenerate into producing low-variance, near-static frames as a conservative strategy to avoid compounding temporal errors. In contrast, by preserving the global layout through sink tokens and the intermediate motion trajectory through the compressed mid-buffer, our framework provides a reliable, uninterrupted spatiotemporal reference. This comprehensive contextual grounding encourages the model to confidently synthesize complex, continuous motion without collapsing into static artifacts.

\noindent \textbf{Limitations and Future Work.}
While \methodname{} excels in generating dynamic content, we observe a nuanced trade-off. Baselines such as LongLive achieve marginally higher Subject Consistency ($92.00$ versus our $90.49$) at the severe expense of motion richness, yielding a Dynamic Degree of only $44.53$ compared to our $56.25$. Closing this consistency gap by further enhancing strict subject preservation without compromising the high dynamic diversity enabled by our compressed history remains a primary direction for future work.




\section{Conclusion}
\label{sec:conclusion}
In this paper, we introduce \textbf{\methodname{}}, a unified framework that fundamentally resolves the dual bottlenecks of error accumulation and unbounded memory growth in autoregressive video generation. By strategically partitioning the KV cache into \textbf{sink}, \textbf{compressed mid}, and \textbf{recent} tokens, our approach strictly bounds the memory footprint to ${\sim}4$\,GB and ensures constant-time attention complexity without discarding essential historical context. Empowered by a $128\times$ dual-branch compression module (${\sim}32\times$ token reduction), incremental RoPE adjustments, and dynamic context selection, \methodname{} achieves a remarkable $24\times$ temporal extrapolation (5\,s $\rightarrow$ 120\,s). This demonstrates that short-video supervision is entirely sufficient for high-quality, long-video synthesis. Ultimately, it generates highly coherent 2-minute videos, establishing state-of-the-art VBench scores and the most robust text-video alignment among existing baselines, paving the way for efficient, unbounded video generation on standard hardware.

%
%
\bibliographystyle{iclr2023_conference}
\bibliography{sample-bibliography}

\clearpage
\onecolumn
\appendix
\startcontents[appendix]
\section*{Appendix Contents}
\printcontents[appendix]{}{1}{\setcounter{tocdepth}{2}}
\clearpage\textbf{}

\section{Extended Discussion on Limitations}
Several directions remain open:
(i)~the fixed compression ratio ($128\times$ volume / ${\sim}32\times$ token) could be made \emph{adaptive} to scene complexity;
(ii)~attention-based importance scoring may not capture all aspects of visual saliency---learned importance predictors could help;
(iii)~scaling to higher resolutions (e.g., $1920{\times}1080$) requires investigating the interaction between spatial compression and quality.
We believe the three-partition principle is general and can be applied to other autoregressive domains beyond video.

\section{Design Comparison of Causal Video Generation Methods}
\begin{table}[h]
\centering
\caption{\textbf{Design comparison} of causal video generation methods. \methodname{} uniquely integrates learned compression, bounded memory, and RoPE adjustment to ensure persistent long-range memory during extended generation.}
\label{tab:method_comparison}

\begin{tabular}{@{}lcccc@{}}
\thickhline
Method & \makecell{KV\\Compress} & \makecell{Bounded\\Memory} & \makecell{RoPE\\Adjustment} & \makecell{Long-range\\Memory} \\
\hline
CausVid~\cite{yin2025slow} & \xmarkg & \xmarkg & --- & \xmarkg \\
Self-Forcing~\cite{huang2025self} & \xmarkg & \xmarkg & --- & \xmarkg \\
Rolling-F.~\cite{liu2025rolling} & \xmarkg & \cmark & \cmark & \xmarkg \\
LongLive~\cite{yang2025longlive} & \xmarkg & \cmark & N/A & partial \\
DeepForcing~\cite{yi2025deep} & \xmarkg & \cmark & \cmark & partial \\
\textbf{PackForcing (Ours)} & \cmark & \cmark & \cmark & \cmark \\
\thickhline
\end{tabular}

\end{table}

\section{Training Strategy}
\label{sec:training}

Our training procedure closely follows the two-phase Self-forcing paradigm~\cite{huang2025self}. First, the causal student model $G_\theta$ is initialized from a pretrained bidirectional prior (e.g., Wan2.1-T2V-1.3B~\cite{wan2025}) via ODE trajectory alignment. Second, the student performs block-wise rollout and is optimized via score distillation against a frozen bidirectional teacher. 

Since the overarching loss formulation and gradient normalization techniques remain identical to standard self-forcing, we omit the standard score-matching equations for brevity. \textbf{The critical distinction in our training pipeline} is the end-to-end optimization of the HR compression layer (Sec.~\ref{sec:compression}). During the rollout phase, the compression module is integrated directly into the computational graph. This joint optimization ensures that the compressed \emph{mid} tokens are explicitly tailored to preserve essential semantic and structural cues for downstream causal attention, rather than merely minimizing a generic pixel-level reconstruction loss.

\noindent \textbf{Short-to-Long Generalization.}
Training uses only 20 latent frames (${\sim}5$\,s after $4\times$ VAE temporal decompression), yet the model generalizes to 2-minute generation---a $24\times$ temporal extrapolation.
This transfer is enabled by the three-partition design: the attention context seen by each transformer layer is bounded at ${\sim}27{,}872$ tokens during \emph{both training and inference}, since compression maintains a constant-size window regardless of the actual video length.
The model therefore never encounters a context distribution it was not trained on.

\section{Streaming VAE Decode}
\label{sec:streaming}

To reduce latency, \methodname{} supports streaming VAE decoding: each block is decoded incrementally as it is generated rather than accumulated for joint decoding.
The VAE decoder maintains a temporal cache to ensure seamless frame boundaries---the first block produces 13 pixel frames (due to the initial receptive field), while subsequent blocks produce 16 each via cache-assisted decoding.
This reduces the time-to-first-frame and enables progressive display.

\section{Additional Experimental Results}
\subsection{Detailed Temporal CLIP Scores}
\label{appendix:clip_scores}

In Table~\ref{tab:ablation_sink_appendix}, we provide the fine-grained, temporal breakdown of the CLIP scores over 20-second intervals for different sink sizes. Consistent with the main text, $N_\text{sink}{=}0$ exhibits a continuous degradation over time (dropping to 28.51 by 120\,s), while $N_\text{sink} \ge 8$ successfully maintains a stable score trajectory throughout the entire minute-scale generation.

\begin{table}[h] 
\centering
\caption{\textbf{Detailed temporal CLIP Score breakdown.} Evaluated across 20-second intervals for different sink sizes. $N_\text{sink} \ge 8$ demonstrates robust prevention of temporal degradation.}
\label{tab:ablation_sink_appendix}

\begin{tabular}{c|cccccc|c}                                                                                                   
\thickhline
Sink Size & 0--20s & 20--40s & 40--60s & 60--80s & 80--100s & 100--120s & Overall \\
\hline
0  & 34.72 & 33.24 & 31.51 & 30.44 & 29.01 & 28.51 & 31.24 \\
2  & 35.32 & 35.45 & 34.64 & 34.24 & 34.63 & 34.80 & 34.85 \\
4  & 35.33 & 35.45 & 34.64 & 34.24 & 34.63 & 34.79 & 34.85 \\
8  & 35.59 & 35.16 & 35.04 & 35.14 & 34.81 & 34.81 & 35.09 \\
16 & \textbf{35.54} & \textbf{35.34} & \textbf{35.21} & \textbf{35.42} & \textbf{35.40} & \textbf{35.46} & \textbf{35.39} \\
\thickhline
\end{tabular}
\end{table}

\section{Inference Algorithm}
\label{sec:algorithm}

Algorithm~\ref{alg:inference} summarizes the complete \methodname{} inference procedure. The key operations involve: (1) block-wise causal generation with multi-step denoising, (2) KV cache updates after each block, (3) periodic dual-branch compression of mid-range frames, and (4) incremental RoPE adjustment for capacity management coupled with streaming VAE decoding for immediate frame output.

\begin{algorithm}[ht]

\caption{\methodname{} Inference}
\label{alg:inference}
\KwIn{Noise $\boldsymbol{\epsilon}$, text prompt $\mathbf{c}$, initial frames $\mathbf{z}_0$}
\KwOut{Generated video frames $\mathbf{V}$}
Initialize KV cache $\mathcal{C} = \{\mathcal{C}_\text{sink}, \mathcal{C}_\text{mid}, \mathcal{C}_\text{recent}\}$\;
Cache initial frames into $\mathcal{C}_\text{sink}$\;
$t_\text{gen} \leftarrow N_\text{sink}$\;
\For{each block $i = 1, \ldots, N_\text{blocks}$}{
  $\mathbf{z}_i \leftarrow \boldsymbol{\epsilon}_i$ \tcp*{Initialize from noise}
  
  \For{$s = 1, \ldots, S$}{  \tcp*{$S{=}4$ denoising steps}
    \If{$s = 1$}{
      $\mathcal{S}_\text{mid} \leftarrow \text{Affinity\_Routing}(\mathcal{C}_\text{mid}, \mathbf{z}_i, N_\text{top})$ \tcp*{Dynamic context selection}
    }
    $\mathcal{C}_\text{active} \leftarrow \{\mathcal{C}_\text{sink}, \mathcal{C}_\text{mid}[\mathcal{S}_\text{mid}], \mathcal{C}_\text{recent}\}$ \tcp*{Sparse attention context}
    
    $\mathbf{v} \leftarrow f_\theta(\mathbf{z}_i, \sigma_s, \mathbf{c}, \mathcal{C}_\text{active})$ \tcp*{Attend to active partitions}
    $\hat{\mathbf{z}}_i \leftarrow \mathbf{z}_i - \sigma_s \cdot \mathbf{v}$\;
    $\mathbf{z}_i \leftarrow (1-\sigma_{s+1})\hat{\mathbf{z}}_i + \sigma_{s+1}\boldsymbol{\epsilon}'$\;
  }
  Update $\mathcal{C}_\text{recent}$ with denoised ${\mathbf{z}}_i$\;
  $\mathbf{V}_i \leftarrow \text{Streaming\_VAE\_Decode}({\mathbf{z}}_i)$ \tcp*{Decode incrementally}
  
  \If{$t_\text{gen} \ge N_\text{sink} + N_\text{mid} + N_\text{rc}$}{ \tcp*{Context window is full}
    $\tilde{\mathbf{h}} \leftarrow \text{HR}({\mathbf{z}}_{i-1}) + \text{LR}(\mathbf{V}_{i-1})$ \tcp*{LR uses streaming decoded output}
    Append compressed tokens to $\mathcal{C}_\text{mid}$\;
    \If{$|\mathcal{C}_\text{mid}| > N_\text{mid}$}{
      Shift $\mathcal{C}_\text{mid}$ window by $\Delta$ blocks \tcp*{Maintain capacity budget}
      Apply incremental RoPE adjustment (Eq.~\ref{eq:rope_correct}) to $\mathcal{C}_\text{sink}$\;
    }
  }
  $t_\text{gen} \leftarrow t_\text{gen} + B_f$\;
}
\Return{Concatenated frames $\mathbf{V}$}
\end{algorithm}

\section{Detailed Mechanism of Dynamic Context Selection}
\label{appendix:dynamic_selection}

In this section, we provide the detailed formulation and engineering optimizations of the Dynamic Context Selection introduced in Sec.~\ref{sec:participative}. Our goal is to dynamically identify the top-$K$ most informative compressed mid-blocks without introducing noticeable latency to the autoregressive generation pipeline.

Given the recent and current blocks acting as queries $Q \in \mathbb{R}^{B \times L_q \times N_h \times d_h}$, and the candidate mid-blocks acting as keys $K \in \mathbb{R}^{M \times B \times L_k \times N_h \times d_h}$ (where $M$ is the number of candidate blocks), we compute an aggregated importance score $s_m$ for each candidate block $m \in \{1, \dots, M\}$.

To minimize the computational overhead of the dot-product attention, we introduce three structural optimizations:
\textbf{(1) Deterministic Query Subsampling:} We uniformly sample a subset of query tokens $\mathcal{S}_q$ from the $L_q$ sequence using a target sampling ratio $\gamma$ (e.g., yielding $\max(32, \lfloor \gamma L_q \rfloor)$ tokens), preserving the spatial-temporal distribution without evaluating the full grid.
\textbf{(2) Half-Head Evaluation:} Since importance is highly correlated across attention heads, we compute the affinity using only the first $N_{opt} = N_h/2$ heads.
\textbf{(3) Step-wise Caching:} The affinity distribution shifts minimally within the internal denoising loop of a single generated chunk. Therefore, we strictly compute $s_m$ only at the \textbf{first denoising timestep} $t=T$, caching the sorted indices for all subsequent $t<T$. Furthermore, an optional interval hyperparameter allows reusing the same indices across multiple adjacent video blocks.

Formally, the importance score $s_m$ for block $m$ is defined as the averaged multi-head affinity over the batch and selected heads, summed across the subsampled queries and full block keys:
\begin{equation}
s_m = \sum_{j=1}^{L_k} \sum_{i \in \mathcal{S}_q} \left( \frac{1}{B \cdot N_{opt}} \sum_{b=1}^{B} \sum_{h=1}^{N_{opt}} \frac{Q_{b, h, i} K_{m, b, h, j}^\top}{\sqrt{d_h}} \right)
\end{equation}

Once $s_m$ is computed for all $M$ candidates, we select the indices of the top-$K$ scores. To maintain the causal temporal structure essential for the Rotary Position Embedding (RoPE) and temporal consistency, the selected indices are sorted in ascending order before retrieving the corresponding Key-Value tensors from the mid-buffer. The selected compact KV caches are then concatenated with the recent blocks for the standard flash-attention forward pass.

\section{Further Analysis}

\subsection{Empirical Analysis of Temporal Attention Patterns}
\label{sec:attention_analysis_app}

A central question in designing KV cache compression for causal video generation is: \emph{which historical tokens does the denoising network actually attend to?} If attention were concentrated on a small, predictable subset of the history---say, only the initial frames and the most recent context---a simple sliding-window or FIFO eviction policy would suffice. However, our empirical analysis reveals a fundamentally different picture.

\paragraph{Setup.}
We generate a 30-second video (480 latent frames, decoded at 16\,fps) using our causal inference pipeline with participatory compression enabled (sink size $=8$, budget $=16$ mid-blocks, block size $=4$ frames). At each compression step, we record the per-block importance score $\phi_j = \sum_i \mathbf{q}_i^\top \mathbf{k}_j / \sqrt{d}$, where queries $\mathbf{q}_i$ are drawn from the recent and current blocks and keys $\mathbf{k}_j$ from each candidate mid-block. We also record which blocks are selected (top-$k$ by importance) for retention. Results are visualized in~\cref{fig:attention_analysis}.

\paragraph{Observation 1: Attention demand spans the full history.}
\cref{fig:attention_analysis}(a) shows the block selection density as a function of relative position in the KV cache. If only sink and recent tokens mattered, we would observe activity exclusively at the left ($\text{position} \approx 0$) and right ($\text{position} \approx 1$) edges. Instead, selected blocks are distributed across the \emph{entire} temporal range, with persistent demand in the mid-range ($0.2$--$0.8$). This is further confirmed by~\cref{fig:attention_analysis}(c), where the average importance score is plotted against relative position for late-stage generation (frame $\geq 200$). The curve is remarkably flat with a mean of $0.499$, indicating near-uniform attention demand over all historical positions. This observation rules out FIFO-based eviction, which would systematically discard early tokens that remain equally relevant.

\paragraph{Observation 2: Important tokens are sparsely and unpredictably distributed.}
\cref{fig:attention_analysis}(b) displays the continuous importance scores for each candidate block across the generation process. High-importance blocks (bright spots) appear at scattered, non-contiguous positions rather than forming coherent temporal clusters. Moreover,~\cref{fig:attention_analysis}(d) quantifies the selection dynamics: the Jaccard distance between consecutive selection sets averages $0.75$, meaning $75\%$ of the retained blocks change at every step. The position diversity metric---the fraction of distinct cache positions visited within a sliding window of 10 steps---stabilizes above $0.85$, confirming that the model rapidly cycles through diverse temporal locations.

\paragraph{Implications for compression design.}
These two observations jointly motivate our compression strategy. First, since attention demand persists across all historical positions, we must retain a representative summary of the \emph{entire} past rather than only recent context---hence the three-region cache structure (sink $+$ mid $+$ recent). Second, since individual mid-range tokens are each only \emph{sporadically} needed while the aggregate demand covers the full history, there is high temporal redundancy among mid-tokens. This makes them ideal candidates for aggressive compression: a compact representation can preserve the distributed information without retaining every individual token. Our dual-shift compression network exploits this redundancy by compressing mid-range KV blocks at a high ratio, achieving constant memory usage regardless of video length while maintaining generation quality.

\begin{table}[t]
\centering 
\caption{Ablation on RoPE Corr.}
\label{tab:ablation_rope}
\begin{tabular}{cccc} 
\toprule 
\makecell{RoPE\\Corr.} & \makecell{CLIP\\(0--20\,s)}\,$\uparrow$ & \makecell{CLIP\\(40--60\,s)}\,$\uparrow$ & $\Delta\,\downarrow$ \\
\midrule 
\xmark & 33.95 & 31.42 & 2.53 \\
\textbf{\cmark} & \textbf{34.02} & \textbf{33.07} & \textbf{0.95} \\
\bottomrule 
\end{tabular}
\end{table}
\subsection{Effect of RoPE Adjustment.}
Table~\ref{tab:ablation_rope} isolates the impact of incremental RoPE rotation.
Without Adjustment, the CLIP gap between early (0--20\,s) and late (40--60\,s) segments is $2.53$---indicating progressive semantic drift after FIFO eviction begins around the 20\,s mark.
With Adjustment, the gap shrinks to $0.95$ (a $62\%$ reduction), confirming that position continuity is essential for stable long-horizon generation.
The cost is negligible: one complex multiplication per eviction event, amounting to $<$0.1\% of total FLOPs.

\subsection{Computational Efficiency Analysis}

Table~\ref{tab:memory} compares memory footprint and speed for 120\,s generation. Without compression, the KV cache alone requires ${\sim}138$\,GB ($748{,}800 \times 30 \times 2 \times 12 \times 128 \times 2$\,bytes), exceeding single-GPU capacity. \methodname{} bounds KV cache memory to ${\sim}4.2$\,GB regardless of video length, while maintaining competitive speed through the reduced attention context.

\begin{table}[t]
\centering
\caption{\textbf{Memory and speed} for 120\,s generation ($832{\times}480$, 16\,FPS, single A100-80GB).}
\label{tab:memory}

\begin{tabular}{@{}lccc@{}}
\toprule
Method & KV Cache & Peak GPU & Speed \\
\midrule
Full cache            & ${\sim}138$\,GB & OOM    & ---       \\
Window-only           & 3.1\,GB        & 24\,GB & 18\,FPS  \\
\methodname{} (FIFO)  & 4.0\,GB        & 26\,GB & 16\,FPS  \\
\methodname{} (Part.) & 4.2\,GB        & 27\,GB & 15\,FPS  \\
\bottomrule
\end{tabular}

\end{table}



In this section, we present extended 120-second qualitative comparisons to further demonstrate the robust long-range consistency and dynamic generation capabilities of PackForcing against state-of-the-art baselines.

\section{More Qualitative Results}
\begin{figure}
    \centering
    \includegraphics[width=1\linewidth]{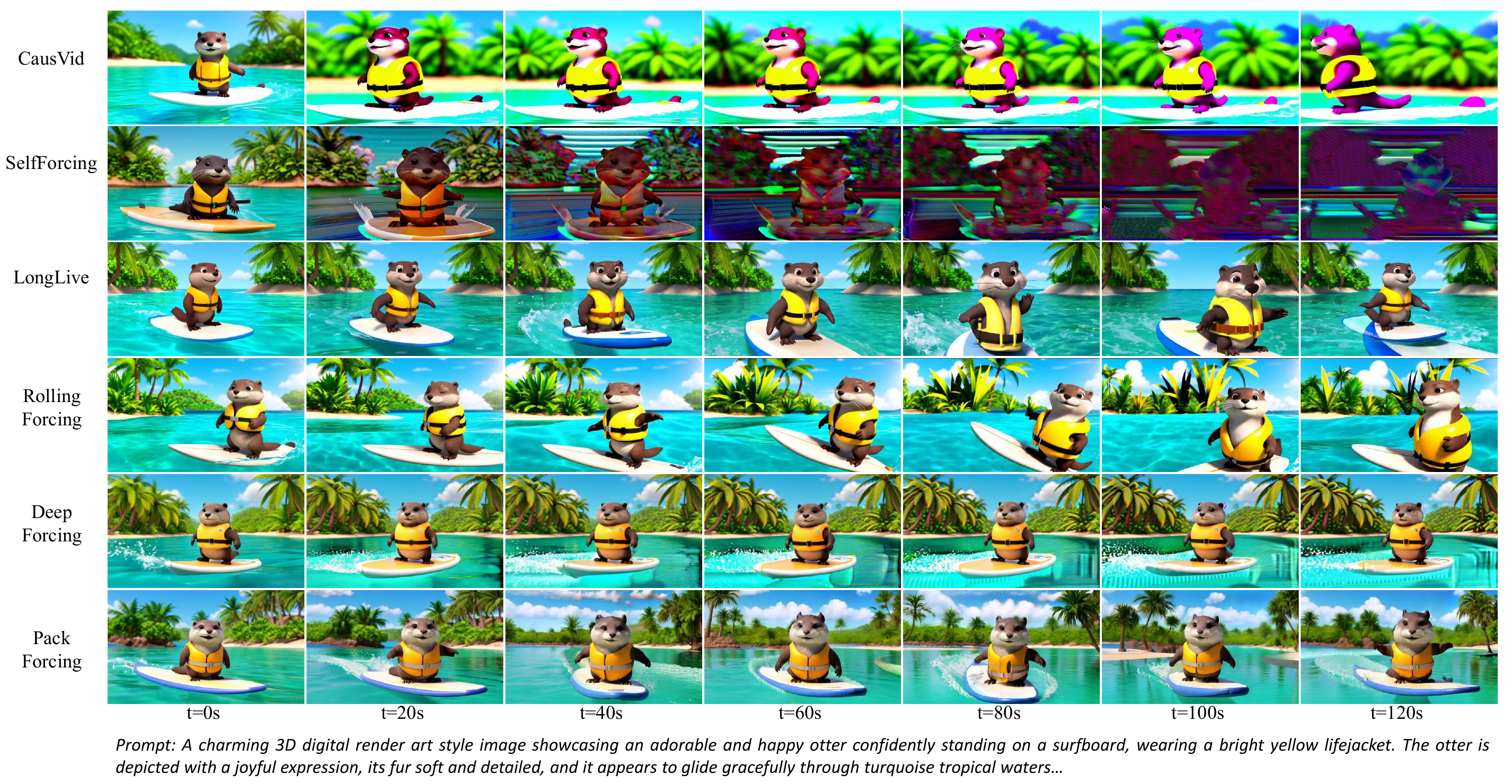}
    \caption{Qualitative Results (1)}
    \label{fig:placeholder1}
\end{figure}
\paragraph{Analysis for Figure 1:}
Figure~\ref{fig:placeholder1} illustrates the generation of a highly dynamic scene featuring an "adorable and happy otter confidently standing on a surfboard". PackForcing successfully maintains strict subject identity, preserving fine details such as the "bright yellow lifejacket" and the "turquoise tropical waters" seamlessly across the entire two-minute duration. In stark contrast, baseline methods struggle with error accumulation over the extended horizon: Self-Forcing exhibits severe color degradation and visual collapse by $t=60s$, while CausVid and Rolling Forcing suffer from noticeable background blurring and subject inconsistency, respectively.

\begin{figure}
    \centering
    \includegraphics[width=1\linewidth]{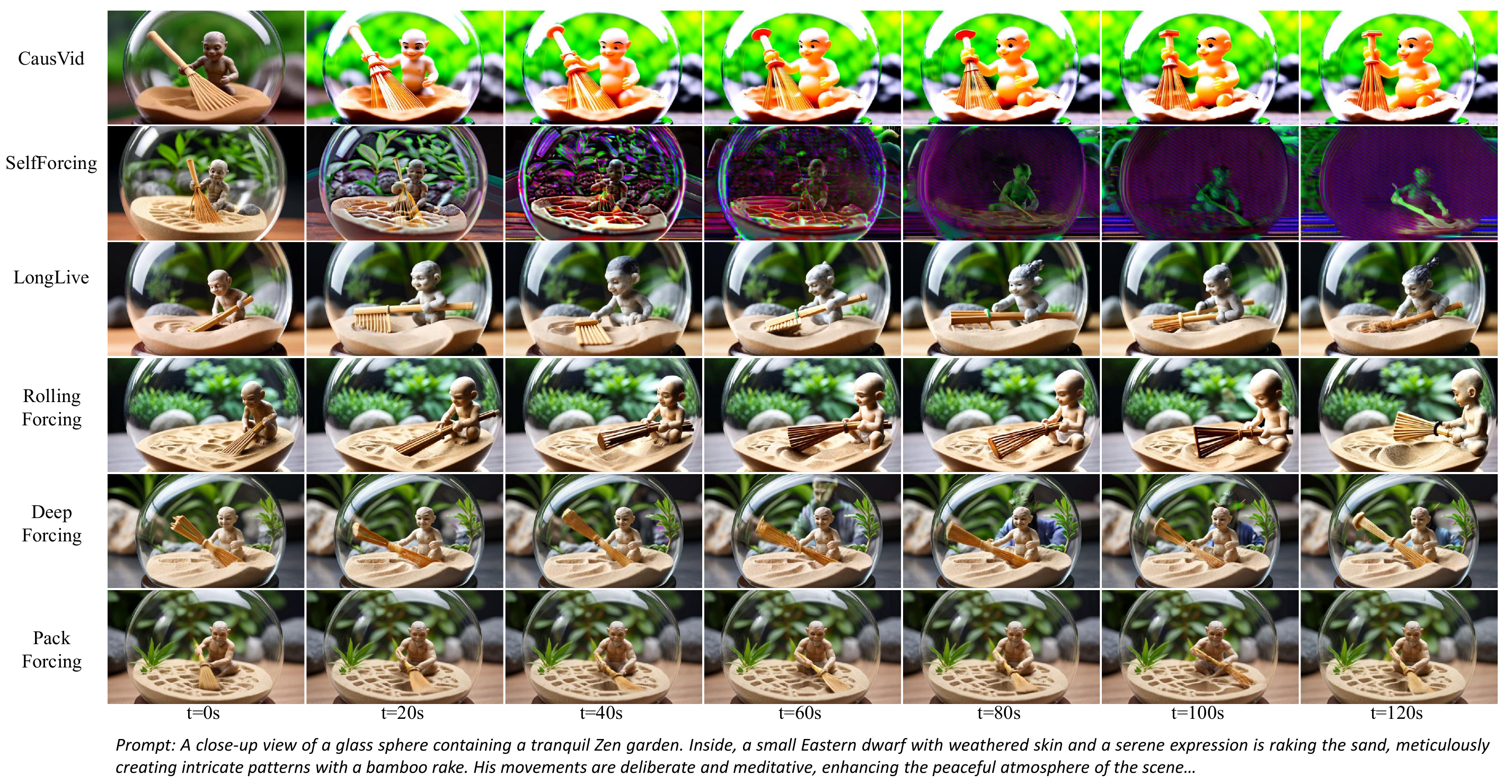}
    \caption{Qualitative Results (2)}
    \label{fig:placeholder2}
\end{figure}

\paragraph{Analysis for Figure 2:}
Figure~\ref{fig:placeholder2} evaluates the models' ability to maintain complex structural boundaries and intricate details, specifically a "glass sphere containing a tranquil Zen garden" and a dwarf holding a "bamboo rake". PackForcing consistently preserves the transparent properties of the glass sphere and the precise identity of the dwarf throughout the 120-second sequence. Baselines fail to sustain this coherence: Self-Forcing completely collapses into dark purple artifacts early in the generation, and DeepForcing gradually loses the structural integrity of the rake and the sphere. While LongLive maintains the subject, it does so at the cost of freezing the motion, whereas our method sustains continuous, deliberate raking movements.

\begin{figure}
    \centering
    \includegraphics[width=1\linewidth]{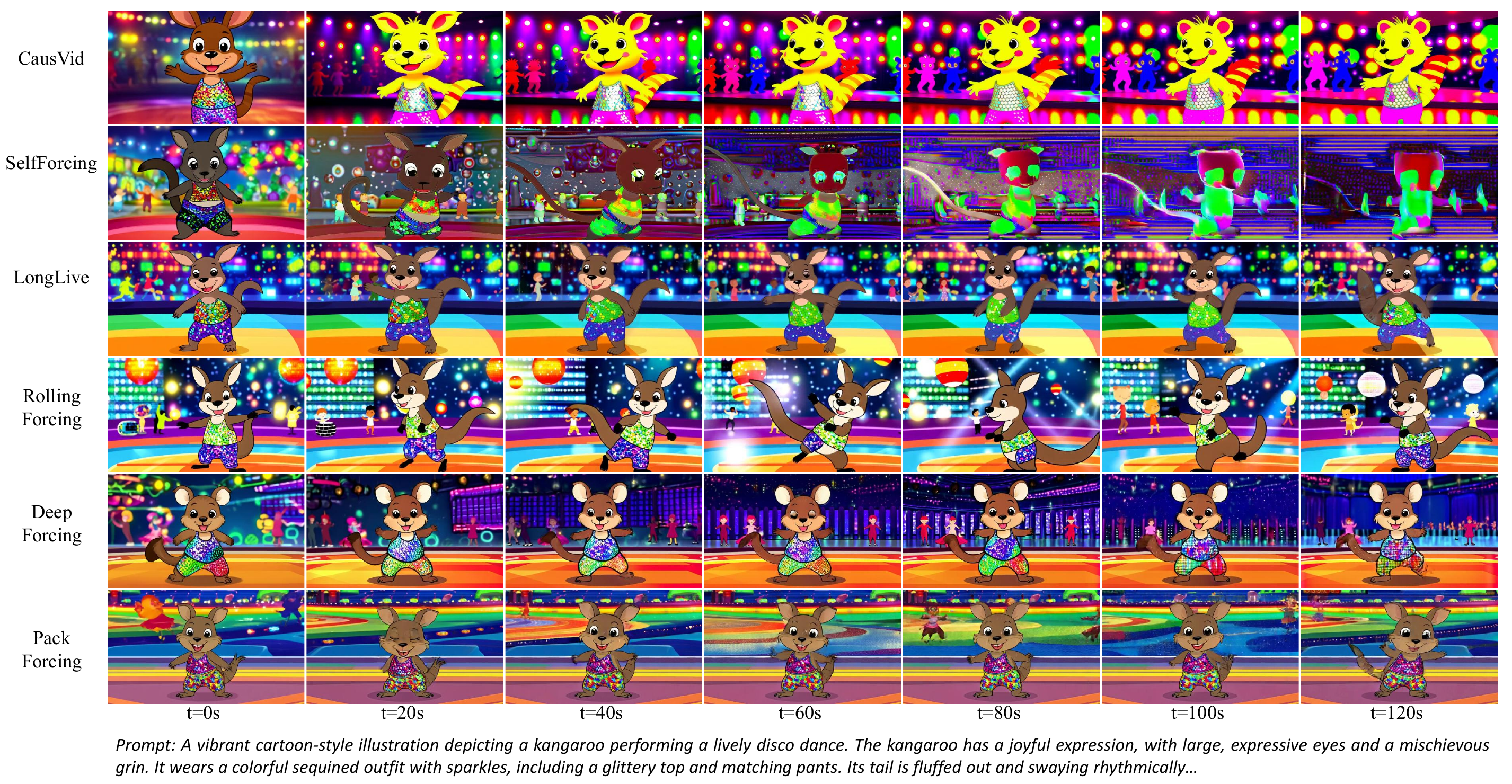}
    \caption{Qualitative Results (3)}
    \label{fig:placeholder3}
\end{figure}
\paragraph{Analysis for Figure 3:}
Figure~\ref{fig:placeholder3} tests the generation of vibrant, stylized content with complex, continuous motion---a "kangaroo performing a lively disco dance" in a "colorful sequined outfit". PackForcing excels in preserving both the high-frequency visual details (the sparkles and vibrant colors) and the highly dynamic rhythmic motion over the extended timeframe. Competing methods typically face a trade-off, either collapsing under the complexity of the prolonged motion or reverting to near-static frames to avoid drift. PackForcing's three-partition memory efficiently prevents these artifacts, ensuring the character remains energetic and visually consistent.